%% file: main.tex
\documentclass{article}
\usepackage{spconf,amsmath,graphicx}
\usepackage{algorithm,algcompatible}
\usepackage{color,epsfig,rotating,subfigure}
\usepackage{amsfonts,amssymb}
\usepackage{subfigure}
\usepackage{algpseudocode}

\title{Pool-based sequential Active Learning with Multi Kernels}%

\name{Jeongmin Chae$^{\dag}$ and Songnam Hong$^{*}$}
\address{$^{\dag}$Electrical Engineering, University of Southern California, CA, USA\\
$^{*}$Electronic Engineering, Hanyang University, Seoul, Korea\\
}

\input{macros}

\DeclareMathOperator*{\argmax}{\arg\!\max}%
\algnewcommand\INPUT{\item[\textbf{Input:}]}%
\algnewcommand\OUTPUT{\item[\textbf{Output:}]}%

\begin{document}
%
\maketitle

\begin{abstract}
We study a pool-based sequential active learning (AL), in which one sample is queried at each time from a large pool of unlabeled data according to a selection criterion. For this framework, we propose two selection criteria, named {\em expected-kernel-discrepancy} (EKD) and {\em expected-kernel-loss} (EKL), by leveraging the particular structure of multiple kernel learning (MKL). Also, it is identified that the proposed EKD and EKL successfully generalize the concepts of popular query-by-committee (QBC) and expected-model-change (EMC), respectively. Via experimental results with real-data sets, we verify the effectiveness of the proposed criteria compared with the existing methods.


\end{abstract}
%


\section{Introduction}\label{sec:intro}

Learning a precise function requires a sufficient number of labeled training samples \cite{scholkopf2001learning,shawe2004kernel,lin2010multiple}. In many applications, however, unlabeled samples are easy to obtain while labeling is expensive or time-consuming. One motivating example is the labeling process of medical data (e.g., magnetic resonance imaging data), since it requires advice by a well-trained expert such as a cardiologist, or a medical imaging expert. The main objective of active learning (AL) is to identify useful data to be labeled by oracle so that a learned function can achieve the best performance as passive learning with the minimum cost. Generally, AL can be classified into  stream-based \cite{bordes2005fast,settles2009active} and  pool-based \cite{sugiyama2009pool,mccallumzy1998employing} methods, where the former makes a decision to query for streaming unlabeled data whereas the latter selects the most informative one to label from a large pool of unlabeled data. 


This paper focuses on pool-based sequential AL problems, in which there is a small set of labeled data and an available large pool of unlabeled data. At each time (or iteration), one sample is selectively queried according to the so-called selection criterion. In this AL framework, thus, a well-designed selection criterion would significantly reduce unnecessary labeling cost. Several selection criterion have been presented in the last decades
\cite{settles2009active,Seung1992,RayChaudhuri1995,DongruiWu2019,DCohn1996,Bdemir2014}. Among them, query-by-committee (QBC) \cite{Seung1992} and expected-model-change (EMC) \cite{RayChaudhuri1995} are the most popular selection criterion. The selection criterion of QBC is determined on the basis of the disagreement among the committee members Also, EMC identifies the unlabeled sample that leads to the largest change on the current model. One motivating example is an expected gradient change approach introduced in \cite{RayChaudhuri1995}, where the change of model is measured by the magnitude of updated gradient of the loss with respect to each sample in a pool.


In this paper we focus on random-feature based multiple-kernel learning (RF-MKL) as the baseline learning method due to its superior performance and scalability \cite{gonen2011multiple,bazerque2013nonparametric,kivinen2004online}. Our major contribution is to develop a pool-based sequential AL suitable for RF-MKL frameworks. Toward this regards, we propose two selection criteria, named {\em expected-kernel-discrepancy} (EKD) and {\em expected-kernel-loss} (EKL). The proposed methods are constructed on the basis of the estimates and reliabilities of $P$ single kernels. Also, we identify that they generalize the concepts of QBC and EMC, by introducing committee members (or models) with possibly different reliabilities. Via experimental results with real-data sets, we demonstrate the effectiveness of the proposed selection criteria over the existing methods.

\vspace{-0.2cm}
\section{Preliminaries}\label{sec:preliminaries}

In this section, we briefly review RF-MKL \cite{rahimi2008random,B.Dai2014} and describe the problem setting of a pool-based sequential AL.

\vspace{-0.2cm}
\subsection{Overview of RF-MKL}\label{subsec:MKL}

Given training data $\{(\xv_{1},y_{1}),...,(\xv_{T},y_{T})\}$, where $\xv_{t} \in \mathcal{X} \subset \RR^{d}$ and $y_{t} \in \mathcal{Y} \subset \RR$, the objective is to train a (non-linear) function $f$, which minimizes the accumulated loss over $T$ training samples.  In RF-MKL frameworks, the learned function of each kernel $i$ has the form of
\begin{equation}\label{eq:function-form}
    \hat{f}_i(\xv)  \triangleq {\hat{\theta}}_{i}^{\trasp}\zv(\xv) \in \Hc_i,\vspace{-0.1cm}
\end{equation}
where $\Hc_i$ represents a reproducing kernel Hilbert space (RKHS) induced by a kernel $\kappa_i$, the optimization variable $\hat{\theta}$ is a 2D-vector, and
\begin{equation}
    \zv(\xv)=\left[\sin{\vv_1^{\trasp}\xv},...,\sin{\vv_{D}^{\trasp}\xv,\cos{\vv_{1}^{\trasp}\xv},...,\cos{\vv_{D}^{\trasp}\xv}}\right]^{\trasp}.\label{eq:z_approx}
\end{equation} Note that the choice of $D$ can determine the accuracy of the RF approximation \cite{rahimi2008random}. 
Then, a learned function is obtained by a convex combination of kernel functions, i.e., \vspace{-0.1cm}
\begin{equation}\label{eq:opt_sol_mkl}
    \hat{f}(\xv)=\sum_{i=1}^{P} \hat{p}_i \hat{f}_i (\xv) \in \bar{\Hc},\vspace{-0.3cm}
\end{equation} where $P$ indicates the number of single kernels, $\bar{\Hc}\eqdef \Hc_1 \bigotimes \Hc_2 \bigotimes \cdots \bigotimes \Hc_P$, and $\hat{p}_i\in[0,1]$ denotes the combination weight of the associated kernel function $\hat{f}_i(\xv)$. The RF-MKL in \eqref{eq:opt_sol_mkl} will be used as the baseline learning framework of the proposed AL. To simplify the notation, let $[N]\eqdef \{1,...,N\}$ for some positive integer $N$.

\subsection{Problem setting}

We consider a learning problem with $M$ unlabeled samples $\mathcal{S}_1=\{\xv_{\tau}: \tau \in [M]\}$, where $\xv_{\tau} \in \mathcal{X} \subset \mathbb{R}^{d}$. Our goal is to choose $T<M$ training samples (or labeled data) actively so that a learned function $\hat{f}(\xv)$ can minimize the generalization error (or test error) of $M-T$ remaining unlabeled data. A pool-based sequential AL is also assumed, in which one sample is queried at each time. Namely, at each time $t$, a selection criterion actively chooses a sample $\xv_{\tau}$ from the pool of unlabeled data, where the set of such unlabeled samples is denoted by $\Sc_{t} \subseteq \Sc_{1}$. In the next section, we will propose new selection criteria especially suitable for RF-MKL frameworks.





\section{Methods}\label{sec:proposed algorithm}

We develop a pool-based sequential AL consisting of two steps: {\em active sample selection} and {\em active function update}. Specifically, at each time $t$, the proposed selection criterion actively selects a sample $\xv_{t}$ on the basis of the latest kernel functions $\{\hat{f}_{t,i}:i\in[P]\}$ and their reliabilities $\{\hat{p}_{t,i}:i\in[P]\}$. Also, they are sequentially updated with the latest labeled data, i.e., $\{\hat{f}_{t+1,i}:i\in[P]\}$ and $\{\hat{p}_{t+1,i}:i\in[P]\}$ are obtained using the $(\xv_{t},y_{t})$. The detailed procedures are provided in the following subsections.

\subsection{The proposed selection criteria}

We explain the proposed selection criteria, called EKD and EKL, by focusing on time (or iteration) $t$. Recall that $\Sc_t$ denotes the set of unlabeled samples at this iteration, i.e., $\Sc_{t} = \Sc_{1} \setminus \Uc_{t-1}$, where $\Uc_{t-1}$ denotes the set of the selected data during the $t-1$ iterations. Then, the most informative sample is determined as
\begin{equation}\label{sel:opt}
\xv_{t}=\argmax_{\xv_\tau \in \Sc_t} \Qc_{t}(\xv_{\tau}),
\end{equation} where $\Qc_{t}$ measures the usefulness of unlabeled samples. Our main goal is to construct the so-called selection criterion $\Qc_{t}(\cdot)$, by leveraging the useful useful information $\{\hat{f}_{t,i}, \hat{p}_{t,i}: i\in[P]\}$ provided by $P$ kernels. The detailed descriptions of EKD and EKL are provided in the below.

{\bf i) Expected-kernel-discrepancy (EKD):} In this criterion, the informativeness is measured by the largest discrepancy (or inconsistency) of the opinions of $P$ kernels. Specifically, it is assumed that the biggest disagreement among the $P$ kernels has the largest impacts on the model update, thus being able to yield the most useful information on the current model. It can be formally defined as
\begin{equation}
\Qc_{t}(\xv_{\tau})=\sum_{i\in[P]} \hat{p}_{t,i}\Lc\left(\hat{f}_{t,i}(\xv_{\tau}), y_{\tau}\right),
\end{equation}  where $\Lc(\cdot,\cdot)$ represents a loss function (e.g., least-square function). Unfortunately, the above criterion is not evaluated as  $y_{\tau}$ is unknown. Instead, we evaluate the above term in average sense, assuming that $Y_{\tau}$ is a random variable which can take the values 
$\{\hat{y}_{i}=\hat{f}_{t,i}(\xv_{\tau}): i\in[P]\}$ with the probability mass function (PMF) $(\hat{p}_{t,1},...,\hat{p}_{t,P})$. Then, the selection criterion of EKD is obtained as
\begin{align}
    \Qc_{t}(\xv_{\tau})&=\EE\left[\sum_{i\in[P]} \hat{p}_{t,i}\Lc\left(\hat{f}_{t,i}(\xv_{\tau}), Y_{\tau}\right)\right]\nonumber\\
    &=\sum_{j\in[P]} \hat{p}_{t,j}\sum_{i\in [P]} \hat{p}_{t,i}\Lc\left(\hat{f}_{t,i}(\xv_{\tau}),\hat{f}_{t,j}(\xv_{\tau})\right).\label{eq:proposed_EKD}
\end{align}
Note that the weights $\hat{p}_{t,i}$'s play a major role in assigning higher risks on the losses of more reliable kernels since they have more influence on the construction of an estimated function.

{\bf ii) Expected-kernel-loss (EKL):} In this criterion, the informativeness is measured by the biggest loss on the current model, where it is assumed that the kernel $i$ can be optimal with the probability $\hat{p}_{t,i}$. Namely, an optimal function at the iteration $t$ is a random variable $F^{\star}$ which can take the values $\hat{f}_{t,i}$ with the probability $\hat{p}_{t,i}$ for $i\in[P]$. Then, EKL criterion can be mathematically defined as
\begin{align}
   \Qc_{t}(\xv_{\tau})&=\EE\left[\Lc\left(\hat{f}_{t}(\xv_{\tau}), F^{\star}(\xv_{\tau}\right) \right]\nonumber\\
   &= \sum_{i\in[P]}\hat{p}_{t,i}\Lc\left(\hat{f}_{t}(\xv_{\tau}), \hat{f}_{t,i}(\xv_{\tau})\right).\label{eq:proposed_EKL}
\end{align} 
The rationale behind this is that the informative sample would raise the largest change on model parameter (e.g. $\thetav$), which naturally leads us to measure the maximum loss value between the current function and updated one with respect to each sample $\xv_{\tau}$.

{\bf Connections with QBC and EMC:} We remark that the proposed criteria can be regarded as some generalizations of the main concepts of QBC and EMC under RF-MKL frameworks. In this case, QBC (or EMC) can naturally take $P$ kernels as $P$ committee members (or $P$ models), instead of using bootstrapping. Using this and from \cite{RayChaudhuri1995}, QBC and EMC can be determined as follows:
\begin{align}
    \mbox{QBC}&=\frac{1}{P}\sum_{j=1}^{P}\left[\hat{f}_{t,j}(\xv_{\tau})-\frac{1}{P}\sum_{i=1}^{P}\hat{f}_{t,i}(\xv_{\tau})\right]^{2}\label{eq:QBC}\\
    \mbox{EMC}&=\frac{1}{P}\sum_{i=1}^{P}[\hat{f}_{t,i}(\xv_{\tau})-\hat{f}_{t}(\xv_{\tau})]^2.\label{eq:EMC}
\end{align} We first show that EMC is a special case of the proposed EKL by assuming $\Lc(x,y)=[x-y]^2$ and equal reliabilities ($\hat{p}_{t,i}=1/P, i\in [P]$). With the same assumptions, the proposed EKD can be simplified as
\begin{equation}
    \frac{1}{P}\sum_{j=1}^{P}\sum_{i=1}^{P}[\hat{f}_{t,j}(\xv_{\tau})-\hat{f}_{t,i}(\xv_{\tau})]^{2} \geq \mbox{QBC in}\; (\ref{eq:QBC}),
\end{equation} where the inequality is due to the fact that $\EE[g(x)] \geq g(\EE[x])$ for a convex function $g$. Thus, one can think that EKD is a generalization of QBC by taking the reliabilities of committee members into account.

\begin{algorithm}
\caption{Pool-based sequential AL with multi kernels}\label{pool-based active learning algorithm}
\begin{algorithmic}[1]
\State {\bf Input:} Kernels $\kappa_{i}, i\in[P]$, parameters $\eta_{l}, \eta_{g}, D > 0$, the number of labeled data $T$, $\Uc_{0}=\phi$, and a pool of unlabeled samples $\mathcal{S}_{1}$.
\State {\bf Output:} A labeled data set  $\Uc_{T}$ and a function $\hat{f}_{T+1}(\xv)$.
\State {\bf Initialization:} $\hat{\thetav}_{1,i} = {\bf 0}$ and $\hat{w}_{1}(i) = 1$ for $i\in [P]$. 
\State {\bf Training iteration:} for $t=1,\cdots,T$ 

\hspace{-0.8cm}$\diamond$  Active sample selection
\begin{itemize}
    \item[$-$] For all $\xv_{\tau} \in \mathcal{S}_{t}$, constructs $\zv_{i}(\xv_{\tau})$ and $\hat{f}_{t,i}(\xv_{\tau})$ using the kernel $\kappa_{i}$ for $i\in[P]$.
    \item[$-$] Computes the degree of disagreement for each sample in $\mathcal{S}_{t}$ via either \eqref{eq:proposed_EKD} or \eqref{eq:proposed_EKL}.
    \item[$-$] Select the sample with the maximum disagreement value (denoted by $\xv_{t})$ and query for its label $y_{t}$.
    \item[$-$] Update $\mathcal{S}_{t+1} = \mathcal{S}_{t} \setminus \{\xv_{t}\}$ and $\Uc_{t}=\Uc_{t-1}\cup\{\xv_{t}\}$.
\end{itemize}

\hspace{-0.8cm}$\diamond$  Sequential function learning with $(\xv_{t},y_{t})$
\begin{itemize}
\item[$-$]Update $\hat{\thetav}_{t+1,i}$ via \eqref{eq:SGDupdate} for $i\in[P]$.
\item[$-$] Set $\hat{f}_{t+1,i}(\xv)= \hat{\thetav}_{t+1,i}^{\trasp}\zv_{i}(\xv)$ for $i \in [P]$.
\item[$-$] Update $\hat{p}_{t+1,i}$ from \eqref{eq:update_p}.
\item[$-$] Update $\hat{f}_{t+1}(\xv) = \sum_{i=1}^{P}\hat{p}_{t+1,i}\hat{f}_{t+1,i}(\xv)$.


\end{itemize}

\end{algorithmic}
\end{algorithm}
%
\subsection{Sequential kernel-function learning}

We will explain how to obtain $\{\hat{f}_{t,i}(\xv): i\in [P]\}$  and $\{\hat{p}_{t,i}:i\in [P]\}$ in a sequential fashion. As noticed before, they are the key factors of the proposed selection criteria. At time $t$, using the selected data $(\xv_{t},y_{t})$ from either EKD or EKL, the proposed kernel-function update consists of the following two steps:

{\bf i) Local step :} This step optimizes each single kernel independently from other kernels, namely,\vspace{-0.2cm}
\begin{equation}\label{eq:local_step}
    \hat{f}_{t+1,i}(\xv) = \hat{\thetav}_{t+1,i}^{\trasp} \zv_{i}(\xv)~ \text{for}~ i\in[P],\vspace{-0.2cm}
\end{equation} where $\hat{\thetav}_{t+1,i}$ is 2$D$-dimensional parameter vector and $\zv_{i}(\xv)$ is defined in \eqref{eq:z_approx}. $\hat{\thetav}_{t+1,i}$ is optimized via stochastic gradient descent (SGD) at every iteration $t$ such as\vspace{-0.2cm}
\begin{equation}\label{eq:SGDupdate}
    \hat{\thetav}_{t+1,i} = \hat{\thetav}_{t,i} - \eta_{l}\nabla\Lc\left(\hat{\thetav}_{t,i}^{\trasp}\zv_{i}(\xv_{t}),y_{t}\right),\vspace{-0.2cm}
\end{equation} where $\nabla\Lc(\hat{\thetav}_{t,i}^{\trasp}\zv_{i}(\xv_{t}),y_{t})$ denotes the gradient of the loss function defined by $(\xv_{t},y_{t})$.

\vspace{0.1cm}
{\bf ii) Global step :} This step combines every single kernel function in \eqref{eq:local_step} with its proper weight $\{\hat{p}_{t+1,i}, i\in[P]\}$ to seek the best $\hat{f}_{t+1}(\xv)$ in \eqref{eq:opt_sol_mkl} approximation. For the weight update of $P$ kernels, {\it exponential strategy} (EXP strategy) \cite{SBubeck2011} is adopted,
\begin{equation}\label{eq:update_p}
\hat{p}_{t+1,i} = \frac{\exp\left(-\eta_{g}\sum_{\tau=1}^{t}\Lc(\hat{f}_{i,\tau}(\xv_\tau),y_\tau)\right)}{\sum_{i=1}^{P}\exp\left(-\eta_{g}\sum_{\tau=1}^{t}\Lc(\hat{f}_{i,\tau}(\xv_\tau),y_\tau)\right)},
\end{equation} with the initial value $\hat{p}_{1,i}=1/P$ for $i\in[P]$.

\begin{table*}
\caption{Comparison of test MSE $\left(\times 10^{-2}\right)$ performance on real data sets in regression tasks. (O) and (S) followed by data name indicate either online learning \eqref{eq:lf1} or supervised learning approach \eqref{eq:lf2} is used to estimate labels for each data set.}
\vspace{0.05cm}
\label{tb:MSEperformance_r}
\begin{center}
\begin{tabular}{l|c|c|c|c|c|c|c|c|c|c}
\hline
Data set& \multicolumn{2}{c|}{Tom(O)}& 
\multicolumn{2}{c|}{Tom(S)}&
\multicolumn{2}{c|}{Air(O)}& \multicolumn{2}{c|}{Air(S)} &\multicolumn{2}{c}{Power(S)}\\ \hline
         & \multicolumn{1}{l|}{20\%} & \multicolumn{1}{l|}{25\%} & 
         \multicolumn{1}{l|}{20\%} & \multicolumn{1}{l|}{25\%} &
         \multicolumn{1}{l|}{20\%} & \multicolumn{1}{l|}{25\%} & \multicolumn{1}{l|}{20\%} & \multicolumn{1}{l|}{25\%} & \multicolumn{1}{l|}{20\%} & \multicolumn{1}{l}{25\%} \\ \hline
Random &\textcolor{blue}{0.28} & \textcolor{blue}{0.2} & \textcolor{blue}{55.2} & \textcolor{blue}{29.6} & \textcolor{blue}{0.55} & \textcolor{blue}{0.47} & \textcolor{blue}{0.3}  & \textcolor{blue}{0.24} & \textcolor{blue}{0.18} & \textcolor{blue}{0.13}  \\ 
\hline
QBC  & 0.04 & 0.26  & 0.15 & 0.11 &  0.41 & 0.42 & 142.7 & 0.38  & 0.12 & 0.12 \\ \hline
EMC & 0.09 & 0.07 & 1.02 & 0.12 &  0.35 & 0.49 &  0.3 & 0.21 & 0.12 & 0.18  \\
\hline
EKL & \textcolor{red}{0.105}  &\textcolor{red}{0.079}  & \textcolor{red}{0.32}  & \textcolor{red}{0.10}  & \textcolor{red}{0.28} &\textcolor{red}{0.41} &  \textcolor{red}{0.35} & \textcolor{red}{0.21} & \textcolor{red}{0.13} & \textcolor{red}{0.15}\\
\hline
EKD & \textcolor{red}{0.084}  &\textcolor{red}{0.031} & \textcolor{red}{0.10} & \textcolor{red}{0.096} & \textcolor{red}{0.34} &\textcolor{red}{0.34} & \textcolor{red}{0.21} & \textcolor{red}{0.22}& \textcolor{red}{0.14}  & \textcolor{red}{0.11}\\
\hline
\end{tabular}
\end{center}
\end{table*}

\subsection{Supervised function learning}

During the training phase, Algorithm 1 provides the set of labeled data $\Uc_{T}=\{(\xv_{t},y_t): t\in[T]$ and an estimated function $\hat{f}_{T+1}(\xv)$. Using them, our learned function $\hat{f}(\xv)$ can be obtained in the following two approaches. 

The first approach follows the conventional supervised learning approach by optimizing kernel parameters $\{\hat{\thetav}_{i}:i\in[P]\}$ using the labeled data $\Uc_{T}$. From (\ref{eq:function-form}), the parameters are obtained as
\begin{equation}
    \hat{\thetav}_{i} = \Zm_{i}^{\dag}\yv,\;\mbox{for } i\in [P],
\end{equation} where $\Zm_{i}$ denotes the $2D \times T$ data matrix whose $j$th column is equal to $\zv_{i}(\xv_{j})$ for $j\in[T]$, and $\yv=(y_1,...,y_T)^{\trasp}$. Also, $\Zm_{i}^{\dag}$ denotes the pseudo-inverse of $\Zm_{i}$. Note that $\Zm_{i}^{\dag}$ exists as $T$ is usually chosen with $T\gg 2D$. Accordingly, the weights (or reliabilities) of kernels are determined as
\begin{align}
    \hat{p}_{i} = \frac{\exp\left(-\eta_{g}\sum_{t=1}^{T}\Lc\left(\hat{\thetav}_{i}^{\trasp}\zv_{i}(\xv_t),y_t\right)\right)}{\sum_{i=1}^{P}\exp\left(-\eta_{g}\sum_{t=1}^{T}\Lc\left(\hat{\thetav}_{i}^{\trasp}\zv_{i}(\xv_t),y_t\right)\right)}.
\end{align} Then, a learned function is obtained as
\begin{equation}\label{eq:lf2}
    \hat{f}(\xv)=\sum_{i\in [P]} \hat{p}_{i}\hat{\thetav}_{i}^{\trasp}\zv_{i}(\xv).
\end{equation} The major drawback of the above approach is an expensive computational complexity to obtain a pseudo-inverse matrix when $T$ becomes a large, i.e., $M$ is large. This problem can be addressed by taking $\hat{f}(\xv)$ as a consequence of active sampling process, i.e., 
\begin{equation}\label{eq:lf1}
    \hat{f}(\xv) = \hat{f}_{T+1}(\xv).
\end{equation} This learning process can be regarded as an online learning \cite{SBubeck2011}. Finally, the learned function $\hat{f}(\xv)$ will be used to estimate the labels of $M-T$ unlabeled data $\xv_{\tau}\in\Sc_{T}$.

\section{Numerical Results}
\label{sec:numerical results}

In this section we show the superiority of the proposed EKD and EKL for regression tasks. It is remarkable that the proposed criteria can be straightforwardly applied to classification tasks, by properly choosing a loss function. As benchmark methods, random sampling, QBC in (\ref{eq:QBC}), and EMC in (\ref{eq:EMC}) are considered. Least-square loss function is assumed. For simulations, we consider the kernel dictionary with 10 Gaussian kernels whose parameters are given as  $\sigma_{i}^{2}=10^{\frac{j-3}{2}}, i\in [10]$. Because of the randomness caused by $\zv(\xv)$ in \eqref{eq:z_approx} and random sampling, the averaged performances over 10 trials are evaluated. We evaluate test errors for various number of labeled data $T$ such as $T=0.2M$, and $0.25M$. Here, the test error is measured by  mean-square-error (MSE) with the $M-T$ unlabeled data.

Table 1 shows the test errors on real data sets obtained from UCI Machine Learning Repository \cite{UCIrepo}. The detailed descriptions of data sets are given as follows:
\begin{itemize}
    \item {\bf Tom's hardware data} ($M=9725$) is acquired from a forum, where each of 96 features represents such as the number of displays and the number of times a content is displayed. The task is to predict the average number of display. The smaller dataset with $M=2000$ (termed Tom(S)) is included to test algorithms.
    \item {\bf Air quality data} ($M=7322$) is obtained from an array of chemical sensors embedded in an air quality sensor. The goal is to predict the concentration of polluting chemicals in the air. The smaller dataset with $M=2000$ (termed Air(S)) is included to test algorithms.
    \item {\bf Power plant data} ($M=2000$) is obtained from combined cycle power plant, which consists of 4 features such as humidity and vacuum. The goal is to determine hourly electrical energy.
\end{itemize}





\vspace{-0.1cm}
We remark that both function-learning methods in \eqref{eq:lf2} and \eqref{eq:lf1} are considered since in particular, the latter is very efficient to deal with large-size data sets. Namely, \eqref{eq:lf2} is used to estimate labels for small Tom(S), Air(S) and Power(S) data sets while \eqref{eq:lf1} is adopted for large Tom(O) and Air(O) data sets. Reflecting the performances of Tom(O) and Tom(S) data, the stability of EKL and EKD is notable, where they perform successfully on both function-learning methods, whereas random sampling poorly degrades. In addition, we observe that the proposed EKL and EKD demonstrate the superior accuracy at almost every $T$ values than random sampling, while QBC and EMC often perform even worse, suggesting the effectiveness of the proposed methods. Comparing with the existing popular methods, the test errors of EKL and EKD show better and more solid performances than QBC and EMC at almost $T$ values. This implies that the proposed selection criteria are indeed proper to select the most informative samples. For the other data sets, it is remarkable that no fixed tendency is observed between the performances of EKD and EKL, rather it depends on data sets. Nonetheless, the proposed EKD and EKL show better performances than the benchmark methods. Similar trends have been observed in other data sets, although only three data sets are included due to the limit of space.

\vspace{-0.1cm}
\section{Conclusion}
\label{sec:Conclusion}
\vspace{-0.1cm}
We investigated a pool-based sequential AL for RF-MKL frameworks. Toward this, we proposed two selection criteria, called {\it expected-kernel-discrepancy} (EKD) and {\it maximum-kernel-loss} (EKL). Via simulation results, we verified that these criteria reveal their accurate sampling abilities compared with the existing famous methods as QBC and EMC. Also, it was shown that EKD and EKL respectively generalize QBC and EMC into RF-MKL frameworks. One interesting future work is a fine integration with clustering methods to reduce the computational cost.


\vfill\pagebreak




\end{document}

%% file: macros.tex
\long\def\comment#1{}

\newfont{\bbb}{msbm10 scaled 700}

\newfont{\bb}{msbm10 scaled 1100}

\newcommand{\RR}{\mbox{\bb R}}

\newcommand{\EE}{\mbox{\bb E}}

\newcommand{\vv}{{\bf v}}
\newcommand{\xv}{{\bf x}}
\newcommand{\yv}{{\bf y}}
\newcommand{\zv}{{\bf z}}

\newcommand{\Zm}{{\bf Z}}

\newcommand{\Hc}{{\cal H}}

\newcommand{\Lc}{{\cal L}}

\newcommand{\Qc}{{\cal Q}}

\newcommand{\Sc}{{\cal S}}

\newcommand{\Uc}{{\cal U}}

\newcommand{\thetav}{\hbox{\boldmath$\theta$}}

\renewcommand{\arg}{{\hbox{arg}}}

\newcommand{\eqdef}{\stackrel{\Delta}{=}}

\newcommand{\trasp}{{\sf T}}
